\documentclass[letterpaper, 10 pt, conference]{ieeeconf}  % Comment this line out if you need a4paper

\IEEEoverridecommandlockouts                              % This command is only needed if 

\usepackage{cite}
\usepackage{amsmath,amssymb,amsfonts}
\usepackage{algorithmic}
\usepackage[graphicx]{realboxes}
\usepackage{textcomp}
\usepackage[table,xcdraw]{xcolor}
\usepackage{multirow}
\usepackage{array}
\usepackage{tikz}
\usepackage{siunitx}
\usepackage{lscape}
\usepackage{adjustbox}
\usepackage{tabularx}
\usepackage{mleftright}
\usepackage{etoolbox}
\usepackage{xcolor}
\usepackage{amsmath}
\usepackage{amssymb}
\usepackage[normalem]{ulem}
\usepackage{tablefootnote}
\usepackage{verbatim}
\usepackage{graphicx}
\usepackage{booktabs}

\title{\LARGE \bf
Subject-Independent Brain-Computer Interface for Decoding High-Level Visual Imagery Tasks}

\author{Dae-Hyeok Lee$^{1}$, Dong-Kyun Han$^{1}$, Sung-Jin Kim$^{2}$, Ji-Hoon Jeong$^{1}$, and Seong-Whan Lee$^{2}$
\thanks{*This research was supported by the Defense Challengeable Future Technology Program of Agency for Defense Development, Republic of Korea.}
\thanks{$^{1}$D.-H. Lee, D.-K. Han, and J.-H. Jeong are with the Department of Brain and Cognitive Engineering, Korea University, Anam-dong, Seongbuk-ku, Seoul 02841, Korea.
        {\tt\small lee\_dh@korea.ac.kr, dk\_han@korea.ac.kr, and jh\_jeong@korea.ac.kr}}%
\thanks{$^{2}$S.-J. Kim and S.-W. Lee are with the Department of Artificial Intelligence, Korea University, Anam-dong, Seongbuk-ku, Seoul 02841, Korea.
        {\tt\small s\_j\_kim@korea.ac.kr and sw.lee@korea.ac.kr}}%
}

\begin{document}

\maketitle
\thispagestyle{empty}
\pagestyle{empty}

%%%%%%%%%%%%%%%%%%%%%%%%%%%%%%%%%%%%%%%%%%%%%%%%%%%%%%%%%%%%%%%%%%%%%%%%%%%%%%%%
\begin{abstract}
Brain-computer interface (BCI) is used for communication between humans and devices by recognizing humans’ status and intention. Communication between humans and a drone using electroencephalogram (EEG) signals is one of the most challenging issues in the BCI domain. In particular, the control of drone swarms (the direction and formation) has more advantages compared to the control of a drone. The visual imagery (VI) paradigm is that subjects visually imagine specific objects or scenes. Reduction of the variability among subjects’ EEG signals is essential for practical BCI-based systems. In this study, we proposed the subepoch-wise feature encoder (SEFE) to improve the performances in the subject-independent tasks by using the VI dataset. This study is the first attempt to demonstrate the possibility of generalization among subjects in the VI-based BCI. We used the leave-one-subject-out cross-validation for evaluating the performances. We obtained higher performances when including our proposed module than excluding our proposed module. The DeepConvNet with SEFE showed the highest performance of 0.72 among six different decoding models. Hence, we demonstrated the feasibility of decoding the VI dataset in the subject-independent task with robust performances by using our proposed module.\\
\end{abstract}
%%%%%%%%%%%%%%%%%%%%%%%%%%%%%%%%%%%%%%%%%%%%%%%%%%%%%%%%%%%%%%%%%%%%%%%%%%%%%%%%
%%%%%%%%%%%%%%%%%%%%%%%%%%%%%%%%%%%%%%%%%%%%%%%%%%%%%%%%%%%%%%%%%%%%%%%%%%%%%%%%
%%%%%%%%%%%%%%%%%%%%%%%%%%%%%%%%%%%%%%%%%%%%%%%%%%%%%%%%%%%%%%%%%%%%%%%%%%%%%%%%
%%%%%%%%%%%%%%%%%%%%%%%%%%%%%%%%%%%%%%%%%%%%%%%%%%%%%%%%%%%%%%%%%%%%%%%%%%%%%%%%
\section{INTRODUCTION}

Brain-computer interface (BCI) has emerged as a technology with great prospects used for communication between humans and devices by recognizing humans' status and intention. The non-invasive BCI is one of the BCI techniques that has many advantages, such as high stability and low cost compared to the invasive BCI\cite{suk2011subject, lee2020continuous, wolpaw2002brain, lee2018high}. Recently, non-invasive BCI systems have been used for controlling external devices \cite{stawicki2017novel, jeong2019trajectory, meng2016noninvasive} or early detecting some diseases \cite{zhang2017hybrid, zhang2019strength}.

One of the most attractive issues is communicating with a drone using the electroencephalogram (EEG) signals with robust performances because the EEG signals can reflect the humans’ intention \cite{lee2020design, yu2019control, wang2018wearable, karavas2017hybrid, karavas2015effect, lafleur2013quadcopter}. Yu \textit{et al}.\cite{yu2019control} presented a development of a practical implementation of a BCI by utilizing low-cost technologies and asynchronous signal processing techniques for EEG signal acquisition and processing to navigate a quadcopter. Especially, the control of drone swarms has been developed by increasing the level of artificial intelligence techniques. The control of drone swarms includes the direction and formation. Karavas \textit{et al}.\cite{karavas2017hybrid} presented the preliminary results of a hybrid brain-machine interface that combined information from the brain and an external device They instructed the subjects to spread-out and fall-in drone swarms consisting of three drones. Koizumi \textit{et al}.\cite{koizumi2019eeg} examined EEG source activity during the visual motion imagery via comparison with the visual motion perception. They instructed the subjects to imagine the movement of a drone in three planes (up/down, left/right, and forward/backward).

EEG-based BCI systems have been developed with various kinds of paradigms which are approximately sorted into the exogenous and endogenous paradigms. Among them, the visual imagery (VI), one of the endogenous BCI paradigms, lets users imagine more intuitively. VI is a paradigm in which the users visually imagine specific objects or scenes. Hence, compared to other BCI paradigms, VI is an important paradigm for the formation control of specific objects. Since the VI can be performed regardless of the complexity of the movement, it is not difficult to perform the task, and thus the data is less contaminated due to low fatigue \cite{pearson2015mental, sousa2017pure, pearson2019human}. Kosmyna \textit{et al}.\cite{kosmyna2018attending} investigated the possibility to build the VI. They instructed the subjects to observe a visual cue of one of two predefined images (a flower or a hammer) and then imagine the same cue. Kwon \textit{et al}.\cite{kwon2020decoding} developed a 3-dimensional BCI training platform and applied it to assist the user in performing more intuitive imagination in the visual motion imagery experiment. They presented statistical evidence that the visual motion imagery has a high correlation between the prefrontal and occipital brain regions. 

A few groups have conducted the analyses for the VI-based EEG data utilizing the convolutional neural network (CNN). Castro \textit{et al}.\cite{castro2020development} investigated the use of deep learning techniques that involve inherent and embedded feature selection and extraction in their hidden layers. Hence, they did not require an explicit and human-intervened selection of features. Their results represented that deep learning-based BCI systems performed significantly better in comparison to the conventional methods.

Most related studies have needed calibration procedures, which require approximately 20-30 min. for data acquisition and model training. Hence, it is essential to reduce the variability among subjects' EEG signals for practical BCI-based systems. The degradation of performance occurs when we apply the trained EEG decoding model using the dataset of a specific subject to the dataset of other subjects. In other words, the optimal model is different for each subject, and this problem leads to the reduction of practicality. In order to increase the practicality of the BCI system, the high variability among subjects needs to be reduced. Kwon \textit{et al}.\cite{kwon2019subject} proposed a subject-independent framework based on the deep CNN using the motor imagery (MI)-based EEG database. They demonstrated that the classification accuracy of their model outperforms that of subject-dependent models. Jeon \textit{et al}.\cite{jeon2019mutual} proposed a novel framework that learns class-relevant and subject-invariant feature representations in an information-theoretic manner, without using adversarial learning. They evaluated the proposed method over two public large dataset (e.g., GIST MI \cite{cho2017eeg} and OpenBMI \cite{lee2019eeg} dataset). However, most of the studies related to the subject-independent BCI are mainly focused on the MI-based BCI. In other words, few studies are in progress on the subject-independent BCI using the VI.

In this study, we acquired EEG signals in the VI experiment which are related to control the drone swarms using our designed experimental paradigm. The classes used in the VI experiment consisted of the most essential formation control in the control of drone swarms (spread-out, fall-in, and hovering). In addition, we proposed the subepoch-wise feature encoder (SEFE) to improve the performances in the subject-independent task. For the VI classification in the subject-independent tasks, we used the leave-one-subject-out (LOSO) cross-validation for evaluating the performances. To the best of our knowledge, this is the first study to generalization among subjects in the VI-based BCI for controlling the formation of drone swarms with robust performances.

The remainder of this article is organized as follows. In Section II, we explain the materials and methods. Details of the experimental results are showed in Sections III. In Section IV, we discuss the analyses. Finally, Section V concludes this article.
%%%%%%%%%%%%%%%%%%%%%%%%%%%%%%%%%%%%%%%%%%%%%%%%%%%%%%%%%%%%%%%%%%%%%%%%%%%%%%%%
\section{MATERIALS AND METHODS}

\subsection{Subjects}

Ten healthy subjects (S1-S10, ten males, aged 25.5 ($\pm 3.1$)) participated in our experiment. The experimental environment and protocols were approved by the Institutional Review Board at Korea University (KUIRB-2020-0318-01). Before the experiment, all subjects were informed about the experimental protocols and consent according to the Declaration of Helsinki. In addition, we informed the subjects to get adequate sleep (over seven hr.) and avoid any alcohol the day before the experiment.
%%%%%%%%%%%%%%%%%%%%%%%%%%%%%%%%%%%%%%%%%%%%%%%%%%%%%%%%%%%%%%%%%%%%%%%%%%%%%%%%
\begin{figure}[t!]
\centering
\scriptsize
\centerline{\includegraphics[width=0.92\columnwidth]{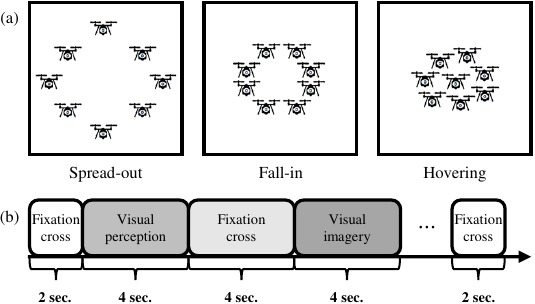}}
\caption{(a) Information of visual cues provided to the subjects in the visual perception section and (b) experimental paradigm for acquiring EEG signals in the VI experiment. The fixation cross of 4 sec. was used for eliminating any possible afterimages.}
\end{figure}   
%%%%%%%%%%%%%%%%%%%%%%%%%%%%%%%%%%%%%%%%%%%%%%%%%%%%%%%%%%%%%%%%%%%%%%%%%%%%%%%%
\subsection{Experimental Environment}

We used the signal amplifier (BrainAmp, Brain Products GmbH, Germany) for measuring EEG signals. We set up the sampling frequency to 500 Hz and a 60 Hz notch filter was used to remove DC noise. We acquired EEG signals using 64 EEG channels that were placed on the subjects’ scalps according to the international 10/20 system. We placed the reference electrode at FCz and the ground electrode at FPz. Before acquiring EEG signals, we injected the conductive gel into the subjects’ scalp to lower the impedance of the EEG electrodes to 10 k$\Omega$ or less.
%%%%%%%%%%%%%%%%%%%%%%%%%%%%%%%%%%%%%%%%%%%%%%%%%%%%%%%%%%%%%%%%%%%%%%%%%%%%%%%%
\begin{figure*}[t!]
\centering
\scriptsize
\centerline{\includegraphics[width=\textwidth]{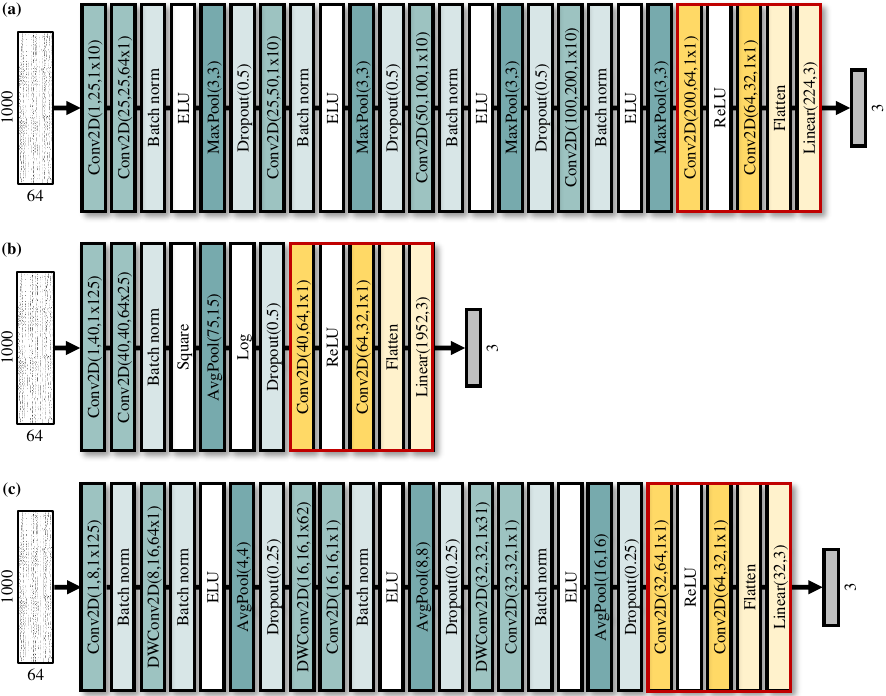}}
\caption{Architectures of three decoding models for EEG classification. (a) DeepConvNet, (b) ShallowConvNet, and (c) EEGNet. The red box indicates the part we applied our proposed SEFE to the three conventional decoding models.}
\end{figure*}   
%%%%%%%%%%%%%%%%%%%%%%%%%%%%%%%%%%%%%%%%%%%%%%%%%%%%%%%%%%%%%%%%%%%%%%%%%%%%%%%%
\subsection{Experimental Paradigm}

The VI paradigm was designed to control the formations of a swarm of drones. Three different classes (spread-out, fall-in, and hovering) were provided to the subjects in the video as shown in Fig. 1(a). One trial consists of four components. A fixation cross is first provided for 2 sec. to measure stable EEG signals. The video is then presented to the subject for 4 sec. in random order among three tasks. A fixation cross is then provided for 4 sec. to remove the afterimage of the video. A blank image is then provided for 4 sec. during which time the subject performs the VI task. We acquired 50 trials per class, and a total of 150 trials were collected per subject. See Fig. 1(b) for more details.

\subsection{EEG Preprocessing}

We conducted the processing of EEG signals using an OpenBMI toolbox \cite {lee2019eeg} and a BBCI toolbox \cite {blankertz2010berlin} in a MATLAB 2019b. The recorded EEG data were band-pass filtered between 0.5 to 50 Hz using a fifth-order Butterworth filter and were downsampled to 250 Hz. 

\subsection{Subepoch-wise feature encoder (SEFE)}

We used three CNN architectures (the DeepConvNet \cite {schirrmeister2017deep}, ShallowConvNet \cite {schirrmeister2017deep}, and EEGNet \cite {lawhern2018eegnet}) as decoding models. The DeepConvNet consists of four convolutional blocks followed by a dense classification layer. Each block consists of convolution, batch norm, an exponential linear unit (ELU) activation function, max pooling, and dropout layer. In particular, the first convolution block consists of consecutive temporal and spatial convolutions, which effectively encode spatial information between EEG channels. The remaining three blocks are constructed using only temporal convolution. The ShallowConvNet is constructed inspired by the filter bank common spatial pattern \cite{ang2008filter}. Like the first convolutional block of the DeepConvNet, it uses consecutive temporal-spatial convolutions, followed by a square activation function, an average pooling, a logarithmic activation function, and finally, a dense classification layer. The last decoding model, the EEGNet, is characterized by a small number of parameters, using depthwise separable convolution. The network consists of temporal convolution and spatial convolution blocks using depth-wise convolution, followed by depth-wise separable convolution blocks. Each convolutional block uses a batch norm and dropout, and an ELU function is used as an activation function.

In this study, we proposed the module that can increase the performances in three conventional models. Our proposed module makes the structure of the classifier deepen relatively by adding the 1$\times$1 convolutional block between the CNN structure of the three decoding models and a dense classification layer as shown in Fig. 2. The 1$\times$1 convolutional block consists of a convolutional layer that has 64 1$\times$1 filters, a rectified linear unit (ReLU) activation function layer, and a convolutional layer that has 32 1$\times$1 filters. Since we instructed the subject to imagine a scene for 4 sec., we assumed that the data features would contain temporal information. Hence, we proposed the SEFE to focus on temporal information when training the model. Here, we have successfully conducted making the classifier deeper while increasing the number of parameters which was a small amount.

\subsection{Performance Evaluation}

We used the LOSO cross-validation for the subject-independent validation. One subject from ten subjects is selected as a test subject. We divided the dataset of the remaining subjects by a ratio of eight (training set) to two (validation set) for each subject. We trained the classifier using the training set. The performance evaluation was then performed using the dataset of the test subject as the input of the classifier.

As described above, the validation set is sampled within the source subjects. We haven’t selected the model based on data from other fully held out subjects that belong to neither the source subjects nor the target subject. Because, unlike other fields that only use the clean dataset, BCI data is very noisy as it is collected at the actual application level, and it also has inefficiency/illiteracy problems \cite{lee2019eeg, wang2020heterogeneous}. When using whole data of a specific subject as a validation set, if the dataset is contaminated with noise or BCI illiterate data, the dataset could not play a role as a complete validation set.

%%%%%%%%%%%%%%%%%%%%%%%%%%%%%%%%%%%%%%%%%%%%%%%%%%%%%%%%%%%%%%%%%%%%%%%%%%%%%%%%
\section{RESULTS}

Fig. 3 indicated an overview of the overall performances of the six different decoding models (three decoding models w/o SEFE and three decoding models w/ SEFE). We obtained higher performances when using the models w/ SEFE than using the models w/o SEFE. When comparing the performances among the models w/ SEFE, the DeepConvNet w/ SEFE showed the highest performance of 0.72, and both ShallowConvNet w/ SEFE and EEGNet w/ SEFE showed a performance of 0.69. In addition, when comparing the performance among the models w/o SEFE, in this case as well, the DeepConvNet w/o SEFE showed the highest performance with 0.67, and the ShallowConvNet w/o SEFE and EEGNet w/o SEFE represented the performance of 0.64 equally.
%%%%%%%%%%%%%%%%%%%%%%%%%%%%%%%%%%%%%%%%%%%%%%%%%%%%%%%%%%%%%%%%%%%%%%%%%%%%%%%%
\begin{figure}[t!]
\centering
\scriptsize
\centerline{\includegraphics[width=0.92\columnwidth]{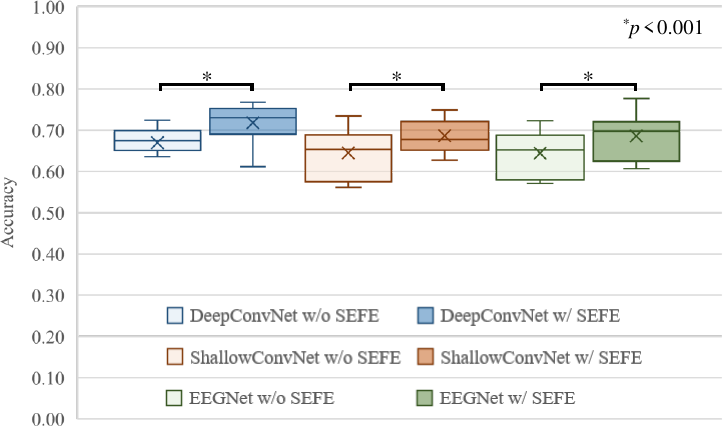}}
\caption{Results of decoding normalized to the average performance across all decoding models including the statistical analyses.}
\end{figure}   
%%%%%%%%%%%%%%%%%%%%%%%%%%%%%%%%%%%%%%%%%%%%%%%%%%%%%%%%%%%%%%%%%%%%%%%%%%%%%%%%

To verify the classification performance difference between the models w/o SEFE and w/ SEFE, we applied the paired \textit{t}‐test with Bonferroni's correction. Initially, we validated the normality and homoscedasticity due to a small number of samples. The normality for each conventional method applying the Shapiro–Wilk test was satisfied with a null hypothesis (H0), and the assumption of homoscedasticity based on Levene’s test was also met for each group. Hence, we conducted a statistical analysis between the models w/o SEFE and w/ SEFE which satisfied these conditions. The statistically significant differences in performance existed between each model w/o SEFE and w/ SEFE (\textit{p}$<$0.001).

%%%%%%%%%%%%%%%%%%%%%%%%%%%%%%%%%%%%%%%%%%%%%%%%%%%%%%%%%%%%%%%%%%%%%%%%%%%%%%%%
\begin{table}[t!]
\caption{Comparison of performances for the VI classification in the subject-independent task between the DeepConvNet w/o SEFE and w/ SEFE}
\renewcommand{\arraystretch}{1.5}
\large
\resizebox{\columnwidth}{!}{
\begin{tabular}{c|ccccc|ccccc}
\hline
\multirow{3}{*}{Subject}           & \multicolumn{5}{c|}{DeepConvNet w/o SEFE \cite{schirrmeister2017deep}}           & \multicolumn{5}{c}{DeepConvNet w/ SEFE} \\ \cline{2-11} 
                          & \textit{\begin{tabular}[c]{@{}c@{}}$1^{st}$\\ acc.\end{tabular}} & \textit{\begin{tabular}[c]{@{}c@{}}$2^{nd}$\\ acc.\end{tabular}} & \textit{\begin{tabular}[c]{@{}c@{}}$3^{rd}$\\ acc.\end{tabular}} & \textit{\begin{tabular}[c]{@{}c@{}}$4^{th}$\\ acc.\end{tabular}} & Average       & \textit{\begin{tabular}[c]{@{}c@{}}$1^{st}$\\ acc.\end{tabular}} & \textit{\begin{tabular}[c]{@{}c@{}}$2^{st}$\\ acc.\end{tabular}} & \textit{\begin{tabular}[c]{@{}c@{}}$3^{rd}$\\ acc.\end{tabular}} & \textit{\begin{tabular}[c]{@{}c@{}}$4^{th}$\\ acc.\end{tabular}} & \begin{tabular}[c]{@{}c@{}}Average       \end{tabular} \\ \hline
S1                        & 0.66                                                        & 0.65                                                        & 0.70                                                        & 0.67                                                        & 0.67          & 0.71                                                        & 0.72                                                        & 0.72                                                        & 0.69                                                        & 0.71                                                   \\
S2                        & 0.70                                                        & 0.68                                                        & 0.71                                                        & 0.70                                                        & 0.70          & 0.76                                                        & 0.73                                                        & 0.75                                                        & 0.75                                                        & 0.75                                                   \\
S3                        & 0.66                                                        & 0.68                                                        & 0.66                                                        & 0.67                                                        & 0.67          & 0.75                                                        & 0.77                                                        & 0.75                                                        & 0.75                                                        & 0.76                                                   \\
S4                        & 0.56                                                        & 0.58                                                        & 0.59                                                        & 0.58                                                        & 0.58          & 0.61                                                        & 0.63                                                        & 0.59                                                        & 0.62                                                        & 0.61                                                   \\
S5                        & 0.71                                                        & 0.68                                                        & 0.71                                                        & 0.70                                                        & 0.70          & 0.76                                                        & 0.74                                                        & 0.76                                                        & 0.75                                                        & 0.75                                                   \\
S6                        & 0.73                                                        & 0.71                                                        & 0.74                                                        & 0.72                                                        & 0.72          & 0.78                                                        & 0.76                                                        & 0.78                                                        & 0.75                                                        & 0.77                                                   \\
S7                        & 0.68                                                        & 0.69                                                        & 0.68                                                        & 0.68                                                        & 0.68          & 0.73                                                        & 0.73                                                        & 0.73                                                        & 0.73                                                        & 0.73                                                   \\
S8                        & 0.69                                                        & 0.69                                                        & 0.68                                                        & 0.69                                                        & 0.69          & 0.74                                                        & 0.72                                                        & 0.72                                                        & 0.74                                                        & 0.73                                                   \\
S9                        & 0.65                                                        & 0.64                                                        & 0.67                                                        & 0.66                                                        & 0.66          & 0.69                                                        & 0.69                                                        & 0.67                                                        & 0.70                                                        & 0.69                                                   \\
S10                       & 0.62                                                        & 0.65                                                        & 0.64                                                        & 0.63                                                        & 0.64          & 0.69                                                        & 0.69                                                        & 0.70                                                        & 0.69                                                        & 0.69                                                   \\ \hline
Average                   & 0.67                                                        & 0.66                                                        & 0.68                                                        & 0.67                                                        & \textbf{0.67} & 0.72                                                        & 0.72                                                        & 0.72                                                        & 0.72                                                        & \textbf{0.72}                                          \\
Std.                      & 0.05                                                        & 0.04                                                        & 0.04                                                        & 0.04                                                        & \textbf{0.04} & 0.05                                                        & 0.04                                                        & 0.05                                                        & 0.04                                                        & \textbf{0.05}   \\ \hline                                    
\end{tabular}}
\end{table}
%%%%%%%%%%%%%%%%%%%%%%%%%%%%%%%%%%%%%%%%%%%%%%%%%%%%%%%%%%%%%%%%%%%%%%%%%%%%%%%%
%%%%%%%%%%%%%%%%%%%%%%%%%%%%%%%%%%%%%%%%%%%%%%%%%%%%%%%%%%%%%%%%%%%%%%%%%%%%%%%%
\begin{table}[t!]
\caption{Comparison of performances for the VI classification in the subject-independent task between the ShallowConvNet w/o SEFE and w/ SEFE}
\renewcommand{\arraystretch}{1.5}
\large
\resizebox{\columnwidth}{!}{
\begin{tabular}{c|ccccc|ccccc}
\hline
\multirow{3}{*}{Subject}           & \multicolumn{5}{c|}{ShallowConvNet w/o SEFE \cite{schirrmeister2017deep}}           & \multicolumn{5}{c}{ShallowConvNet w/ SEFE} \\ \cline{2-11} 
                          & \textit{\begin{tabular}[c]{@{}c@{}}$1^{st}$\\ acc.\end{tabular}} & \textit{\begin{tabular}[c]{@{}c@{}}$2^{nd}$\\ acc.\end{tabular}} & \textit{\begin{tabular}[c]{@{}c@{}}$3^{rd}$\\ acc.\end{tabular}} & \textit{\begin{tabular}[c]{@{}c@{}}$4^{th}$\\ acc.\end{tabular}} & Average       & \textit{\begin{tabular}[c]{@{}c@{}}$1^{st}$\\ acc.\end{tabular}} & \textit{\begin{tabular}[c]{@{}c@{}}$2^{st}$\\ acc.\end{tabular}} & \textit{\begin{tabular}[c]{@{}c@{}}$3^{rd}$\\ acc.\end{tabular}} & \textit{\begin{tabular}[c]{@{}c@{}}$4^{th}$\\ acc.\end{tabular}} & \begin{tabular}[c]{@{}c@{}}Average       \end{tabular} \\ \hline
S1                        & 0.62                                                        & 0.65                                                        & 0.65                                                        & 0.64                                                        & 0.64          & 0.67                                                        & 0.68                                                        & 0.65                                                        & 0.67                                                        & 0.67                                                   \\
S2                        & 0.68                                                        & 0.68                                                        & 0.67                                                        & 0.68                                                        & 0.68          & 0.71                                                        & 0.70                                                        & 0.70                                                        & 0.71                                                        & 0.71                                                   \\
S3                        & 0.65                                                        & 0.65                                                        & 0.67                                                        & 0.65                                                        & 0.66          & 0.66                                                        & 0.66                                                        & 0.64                                                        & 0.65                                                        & 0.65                                                   \\
S4                        & 0.57                                                        & 0.58                                                        & 0.54                                                        & 0.56                                                        & 0.56          & 0.67                                                        & 0.69                                                        & 0.67                                                        & 0.66                                                        & 0.67                                                   \\
S5                        & 0.71                                                        & 0.69                                                        & 0.67                                                        & 0.69                                                        & 0.69          & 0.75                                                        & 0.74                                                        & 0.75                                                        & 0.75                                                        & 0.75                                                   \\
S6                        & 0.74                                                        & 0.74                                                        & 0.74                                                        & 0.72                                                        & 0.74          & 0.76                                                        & 0.76                                                        & 0.73                                                        & 0.75                                                        & 0.75                                                   \\
S7                        & 0.65                                                        & 0.64                                                        & 0.65                                                        & 0.66                                                        & 0.65          & 0.72                                                        & 0.72                                                        & 0.70                                                        & 0.71                                                        & 0.71                                                   \\
S8                        & 0.70                                                        & 0.67                                                        & 0.69                                                        & 0.69                                                        & 0.69          & 0.68                                                        & 0.70                                                        & 0.68                                                        & 0.67                                                        & 0.68                                                   \\
S9                        & 0.57                                                        & 0.58                                                        & 0.57                                                        & 0.57                                                        & 0.57          & 0.64                                                        & 0.66                                                        & 0.66                                                        & 0.64                                                        & 0.65                                                   \\
S10                       & 0.58                                                        & 0.58                                                        & 0.58                                                        & 0.57                                                        & 0.58          & 0.63                                                        & 0.63                                                        & 0.63                                                        & 0.62                                                        & 0.63                                                   \\ \hline
Average                   & 0.65                                                        & 0.65                                                        & 0.64                                                        & 0.64                                                        & \textbf{0.64} & 0.69                                                        & 0.69                                                        & 0.68                                                        & 0.68                                                        & \textbf{0.69}                                          \\
Std.                      & 0.06                                                        & 0.06                                                        & 0.06                                                        & 0.05                                                        & \textbf{0.06} & 0.04                                                        & 0.04                                                        & 0.04                                                        & 0.04                                                        & \textbf{0.04}  \\ \hline                                       
\end{tabular}}
\end{table}
%%%%%%%%%%%%%%%%%%%%%%%%%%%%%%%%%%%%%%%%%%%%%%%%%%%%%%%%%%%%%%%%%%%%%%%%%%%%%%%%
%%%%%%%%%%%%%%%%%%%%%%%%%%%%%%%%%%%%%%%%%%%%%%%%%%%%%%%%%%%%%%%%%%%%%%%%%%%%%%%%
\begin{table}[t!]
\caption{Comparison of performances for the VI classification in the subject-independent task between the EEGNet w/o SEFE and w/ SEFE}
\renewcommand{\arraystretch}{1.5}
\large
\resizebox{\columnwidth}{!}{
\begin{tabular}{c|ccccc|ccccc}
\hline
\multirow{3}{*}{Subject}           & \multicolumn{5}{c|}{EEGNet w/o SEFE \cite {lawhern2018eegnet}}           & \multicolumn{5}{c}{EEGNet w/ SEFE} \\ \cline{2-11} 
                          & \textit{\begin{tabular}[c]{@{}c@{}}$1^{st}$\\ acc.\end{tabular}} & \textit{\begin{tabular}[c]{@{}c@{}}$2^{nd}$\\ acc.\end{tabular}} & \textit{\begin{tabular}[c]{@{}c@{}}$3^{rd}$\\ acc.\end{tabular}} & \textit{\begin{tabular}[c]{@{}c@{}}$4^{th}$\\ acc.\end{tabular}} & Average       & \textit{\begin{tabular}[c]{@{}c@{}}$1^{st}$\\ acc.\end{tabular}} & \textit{\begin{tabular}[c]{@{}c@{}}$2^{st}$\\ acc.\end{tabular}} & \textit{\begin{tabular}[c]{@{}c@{}}$3^{rd}$\\ acc.\end{tabular}} & \textit{\begin{tabular}[c]{@{}c@{}}$4^{th}$\\ acc.\end{tabular}} & \begin{tabular}[c]{@{}c@{}}Average       \end{tabular} \\ \hline
S1                        & 0.64                                                        & 0.68                                                        & 0.67                                                        & 0.65                                                        & 0.66          & 0.69                                                        & 0.69                                                        & 0.71                                                        & 0.70                                                        & 0.70                                                   \\
S2                        & 0.67                                                        & 0.70                                                        & 0.69                                                        & 0.68                                                        & 0.69          & 0.72                                                        & 0.70                                                        & 0.72                                                        & 0.72                                                        & 0.72                                                   \\
S3                        & 0.66                                                        & 0.63                                                        & 0.64                                                        & 0.65                                                        & 0.64          & 0.69                                                        & 0.70                                                        & 0.69                                                        & 0.71                                                        & 0.70                                                   \\
S4                        & 0.55                                                        & 0.58                                                        & 0.57                                                        & 0.58                                                        & 0.57          & 0.62                                                        & 0.60                                                        & 0.62                                                        & 0.65                                                        & 0.62                                                   \\
S5                        & 0.71                                                        & 0.68                                                        & 0.69                                                        & 0.69                                                        & 0.69          & 0.73                                                        & 0.75                                                        & 0.74                                                        & 0.73                                                        & 0.74                                                   \\
S6                        & 0.72                                                        & 0.71                                                        & 0.72                                                        & 0.74                                                        & 0.72          & 0.78                                                        & 0.77                                                        & 0.78                                                        & 0.78                                                        & 0.78                                                   \\
S7                        & 0.61                                                        & 0.66                                                        & 0.63                                                        & 0.64                                                        & 0.64          & 0.68                                                        & 0.68                                                        & 0.70                                                        & 0.66                                                        & 0.68                                                   \\
S8                        & 0.68                                                        & 0.66                                                        & 0.68                                                        & 0.67                                                        & 0.67          & 0.71                                                        & 0.70                                                        & 0.71                                                        & 0.69                                                        & 0.70                                                   \\
S9                        & 0.58                                                        & 0.58                                                        & 0.58                                                        & 0.58                                                        & 0.58          & 0.62                                                        & 0.60                                                        & 0.62                                                        & 0.59                                                        & 0.61                                                   \\
S10                       & 0.57                                                        & 0.56                                                        & 0.61                                                        & 0.58                                                        & 0.58          & 0.62                                                        & 0.62                                                        & 0.63                                                        & 0.64                                                        & 0.63                                                   \\ \hline
Average                   & 0.64                                                        & 0.64                                                        & 0.65                                                        & 0.65                                                        & \textbf{0.64} & 0.69                                                        & 0.68                                                        & 0.69                                                        & 0.69                                                        & \textbf{0.69}                                          \\
Std.                      & 0.06                                                        & 0.05                                                        & 0.05                                                        & 0.06                                                        & \textbf{0.05} & 0.06                                                        & 0.06                                                        & 0.05                                                        & 0.05                                                        & \textbf{0.05}    \\ \hline                                     
\end{tabular}}
\end{table}
%%%%%%%%%%%%%%%%%%%%%%%%%%%%%%%%%%%%%%%%%%%%%%%%%%%%%%%%%%%%%%%%%%%%%%%%%%%%%%%%
%%%%%%%%%%%%%%%%%%%%%%%%%%%%%%%%%%%%%%%%%%%%%%%%%%%%%%%%%%%%%%%%%%%%%%%%%%%%%%%%
\begin{table}[t!]
\centering
\caption{Comparison of the number of parameters among all decoding models}
\renewcommand{\arraystretch}{1.4}
\begin{tabular}{cc}
\hline
Name of model           & \# of parameter  \\ \hline
DeepConvNet w/o SEFE    & 108,485          \\
DeepConvNet w/ SEFE     & \textbf{318,669} \\
ShallowConvNet w/o SEFE & 103,520          \\
ShallowConvNet w/ SEFE  & \textbf{108,224} \\
EEGNet w/o SEFE         & 3,400            \\
EEGNet w/ SEFE          & \textbf{9,832}   \\ \hline
\end{tabular}
\end{table}
%%%%%%%%%%%%%%%%%%%%%%%%%%%%%%%%%%%%%%%%%%%%%%%%%%%%%%%%%%%%%%%%%%%%%%%%%%%%%%%%

Table I, Table II, and Table III showed the overall VI classification performances in the subject-independent task using the decoding models w/o SEFE and w/ SEFE among all subjects. We repeated the validation four times after adopting a different shuffle order each time. In Table I, the DeepConvNet w/ SEFE showed higher average accuracy than the DeepConvNet w/o SEFE, and the numerical difference was 0.05. In the case of two models, S6 showed the highest accuracies of 0.72 and 0.77, respectively, and S4 represented the lowest accuracies of 0.58 and 0.61, respectively. As shown in Table II, the average accuracy of the ShallowConvNet w/ SEFE was 0.05 higher than that of the ShallowConvNet w/o SEFE. In the case of the ShallowConvNet w/o SEFE, the highest accuracy was obtained by S6, but S4 represented the lowest accuracy. Also, in the performance of the ShallowConvNet w/ SEFE, S5 and S6 showed the highest accuracy of 0.75 equally, and S10 represented the lowest accuracy of 0.63. In Table III, the EEGNet w/ SEFE showed 0.05 higher average accuracy than the EEGNet w/o SEFE. In the case of the two models, S6 showed the highest accuracies of 0.72 and 0.78, respectively. Also, S4 and S9 represented the lowest accuracies of 0.57 and 0.61, respectively.

Also, Table IV showed the number of parameters in each model. The differences in the number of parameters existed between each model w/o SEFE and w/ SEFE. We confirmed that the number of parameters increased in all models when including the SEFE. When comparing with the models between including the SEFE and excluding the SEFE, the number of parameters and the performance increased in the case of including the SEFE. In other words, we proved that the increase in the number of parameters due to the SEFE was closely related to performance improvement.
%%%%%%%%%%%%%%%%%%%%%%%%%%%%%%%%%%%%%%%%%%%%%%%%%%%%%%%%%%%%%%%%%%%%%%%%%%%%%%%%
\section{DISCUSSION}

As represented in Fig. 3, Table I, Table II, and Table III, we obtained the highest performance when using the DeepConvNet w/ SEFE among six decoding models. In addition, three models w/ SEFE outperformed three models w/o SEFE. We showed the increase of performances by adding the 1$\times$1 convolutional block in the end part of models. Therefore, the DeepConvNet w/ SEFE that had the deepest structure and the largest number of parameters of 318,669 could show the best performance. In contrast, the EEGNet w/o SEFE which had the least number of parameters of 3,400 showed the lowest performance. In other words, our proposed SEFE increased the number of parameters closely with the performance.

Simply, a large number of parameters does not mean that the performance would be high. The number of parameters in the DeepConvNet w/o SEFE was 108,485, which was larger than those of the ShallowConvNet w/ SEFE and the EEGNet w/ SEFE, but the performances of the ShallowConvNet w/SEFE and the EEGNet w/SEFE were higher than that of the DeepConvNet w/o SEFE. In other words, our proposed SEFE increased the number of parameters while maintaining the inherent characteristics of each model.

For increasing the practicality of BCI, studies for a subject-independent BCI system are essential. When common features are extracted from various source subjects and a classifier is trained using common features, a training session would not be performed from a target subject. However, most subject-independent studies are generally conducted applying the MI. Compared to the MI, the VI has the advantage that subjects have less difficulty in performing tasks, and thus data contamination by fatigue is less. Hence, the VI is an important element in the endogenous BCI system, and a study for the VI-based subject-independent BCI system is essential for the practicality of BCI.

We used a relatively larger amount of data when training the model through the subject-independent model training compared with the subject-dependent model. Since it was essential that the model must capture variability in the source data and extract robust features, the models including the SEFE were more effective in the subject-independent task compared to the shallow models that showed high performance in the conventional subject-dependent task.
%%%%%%%%%%%%%%%%%%%%%%%%%%%%%%%%%%%%%%%%%%%%%%%%%%%%%%%%%%%%%%%%%%%%%%%%%%%%%%%%
\section{CONCLUSION AND FUTURE WORKS}

In this study, we represented the feasibility of decoding the VI-based three classes in the subject-independent task. We obtained EEG signals in the VI experiment for controlling the essential formations of drone swarms (spread-out, fall-in, and hovering). Moreover, we proposed an encoder that makes the structure of the classifier deepen relatively to focus on temporal information when training the model for improving the performances of the VI classification in the subject-independent task. We validated the performances using six decoding models (the DeepConvNet w/o SEFE and w/ SEFE, the ShallowConvNet w/o SEFE and w/ SEFE, and the EEGNet w/o SEFE and w/ SEFE). We showed an increase in performances when using deeper structure models which apply our proposed module to the end part of the conventional models. The DeepConvNet w/ SEFE showed the highest performance of 0.72 among six decoding models.

In future works, we will propose a deep learning-based model that can be used in analyzing the EEG signals which are acquired in various endogenous paradigms for the practical BCI systems and can perform the subject-independent tasks with robust classification performances. To this end, we plan to collect EEG signals based on new experimental paradigms, and in addition, we will apply various data augmentation methods to solve a lack of data problem.
%%%%%%%%%%%%%%%%%%%%%%%%%%%%%%%%%%%%%%%%%%%%%%%%%%%%%%%%%%%%%%%%%%%%%%%%%%%%%%%%
\section*{ACKNOWLEDGMENT}
The authors would like to thank H.-J. Ahn for designing the experimental paradigm and acquiring the EEG data.
%%%%%%%%%%%%%%%%%%%%%%%%%%%%%%%%%%%%%%%%%%%%%%%%%%%%%%%%%%%%%%%%%%%%%%%%%%%%%%%%

\bibliographystyle{IEEEtran}
\bibliography{REFERENCE}

\end{document}